\pdfoutput=1

\documentclass[11pt]{article}

\usepackage[preprint]{acl}

\usepackage{times}
\usepackage{latexsym}
\usepackage[T1]{fontenc}

\usepackage[utf8]{inputenc}

\usepackage{microtype}

\usepackage{inconsolata}

\usepackage{graphicx}
\usepackage{amsmath}
\title{The Tonogenesis Continuum in Tibetan: A Computational Investigation}

\author{Siyu Liang \and Zhaxi Zerong \\
University of Washington \\
\texttt{liangsy, tashi0@uw.edu} \\}

\begin{document}
\maketitle
\begin{abstract}
Tonogenesis—the historical process by which segmental contrasts evolve into lexical tone—has traditionally been studied through comparative reconstruction and acoustic phonetics. We introduce a computational approach that quantifies the functional role of pitch at different stages of this sound change by measuring how pitch manipulation affects automatic speech recognition (ASR) performance. Through analysis on the sensitivity to pitch-flattening from a set of closely related Tibetan languages, we find evidence of a tonogenesis continuum: atonal Amdo dialects tolerate pitch removal the most, while fully tonal Ü-Tsang varieties show severe degradation, and intermediate Kham dialects fall measurably between these extremes. These gradient effects demonstrate how ASR models implicitly learn the shifting functional load of pitch as languages transition from consonant-based to tone-based lexical contrasts. Our findings show that computational methods can capture fine-grained stages of sound change and suggest that traditional functional load metrics, based solely on minimal pairs, may overestimate pitch dependence in transitional systems where segmental and suprasegmental cues remain phonetically intertwined.

\end{abstract}

\section{Introduction}
\label{sec:intro}

Tonogenesis refers to the emergence of lexical tone from earlier consonantal or laryngeal contrasts such as voicing and aspiration \citep{hombert_development_1977, haudricourt_lorigine_1954}. Over time, secondary pitch perturbations can become the primary cue for distinguishing words, turning a formerly atonal language into a tonal one. This process unfolds along a continuum: within a language family or group, some varieties maintain complex onset clusters and show little reliance on pitch (e.g. varieties of Amdo Tibetan), whereas others develop robust tone contrasts (e.g., Lhasa Tibetan) \citep{sun_variegated_2015}. Fully tonal languages (at later stages of tonogenesis) exhibit a strong dependence on fundamental frequency ($f_0$) for lexical contrasts, which can pose significant challenges for automatic speech recognition (ASR) if pitch information is removed \citep{zhang_role_2020, fu_importance_1998}. 

Recent work has established that pitch-flattening—systematically removing $f_0$ contours while preserving spectral information—provides a computational method for quantifying tonal dependence in ASR systems \citep{liang_tone_2025}. Across typologically diverse languages, fully tonal systems suffer dramatic performance degradation when pitch is removed, while non-tonal languages show minimal impact. Building on this validated methodology, we extend the investigation to multiple Tibetan languages representing distinct stages of tonogenesis. We hypothesize that flattening $f_0$ contours will cause a larger performance drop for languages with established tonal contrasts than for those that rely more on consonantal cues. By quantifying the impact of pitch loss on ASR using word error rate and character error rate, we offer computational evidence that captures more granular information than text-based functional load measures in the context of tonogenesis. Our findings demonstrate that dialects at different points in the tonogenesis continuum exhibit correspondingly distinct degrees of pitch dependence, providing new empirical support for understanding how languages transition from consonant-based to tone-based lexical systems.

\section{Literature Review}
\label{sec:lit_review}

\subsection{Tone and tonogenesis}
Tonal languages use pitch ($f_0$) as a primary cue for lexical contrasts, as exemplified by languages like Cantonese, where different pitch patterns on the same segment sequence produce entirely distinct words \citep{yip_tone_2002, matthews_cantonese_2013}. In contrast, non-tonal languages like English employ pitch mainly for intonation and stress \citep{edwards_articulatory_1988}. Crucially, tonogenesis can arise when older segmental cues (e.g., voicing or coda consonants) induce pitch perturbations that eventually become the main phonemic signal for lexical differentiation \citep{haudricourt_lorigine_1954, hombert_phonetic_1979}. In Vietnamese, for instance, the loss of final stops led to newly phonologized tones \citep{thurgood_vietnamese_2002}, while other Southeast Asian languages display partial or incomplete shifts \citep{matisoff_tonogenesis_1973}.

\subsection{Tibetan languages}
\begin{figure}[ht]
    \centering
    \includegraphics[width=0.45\textwidth]{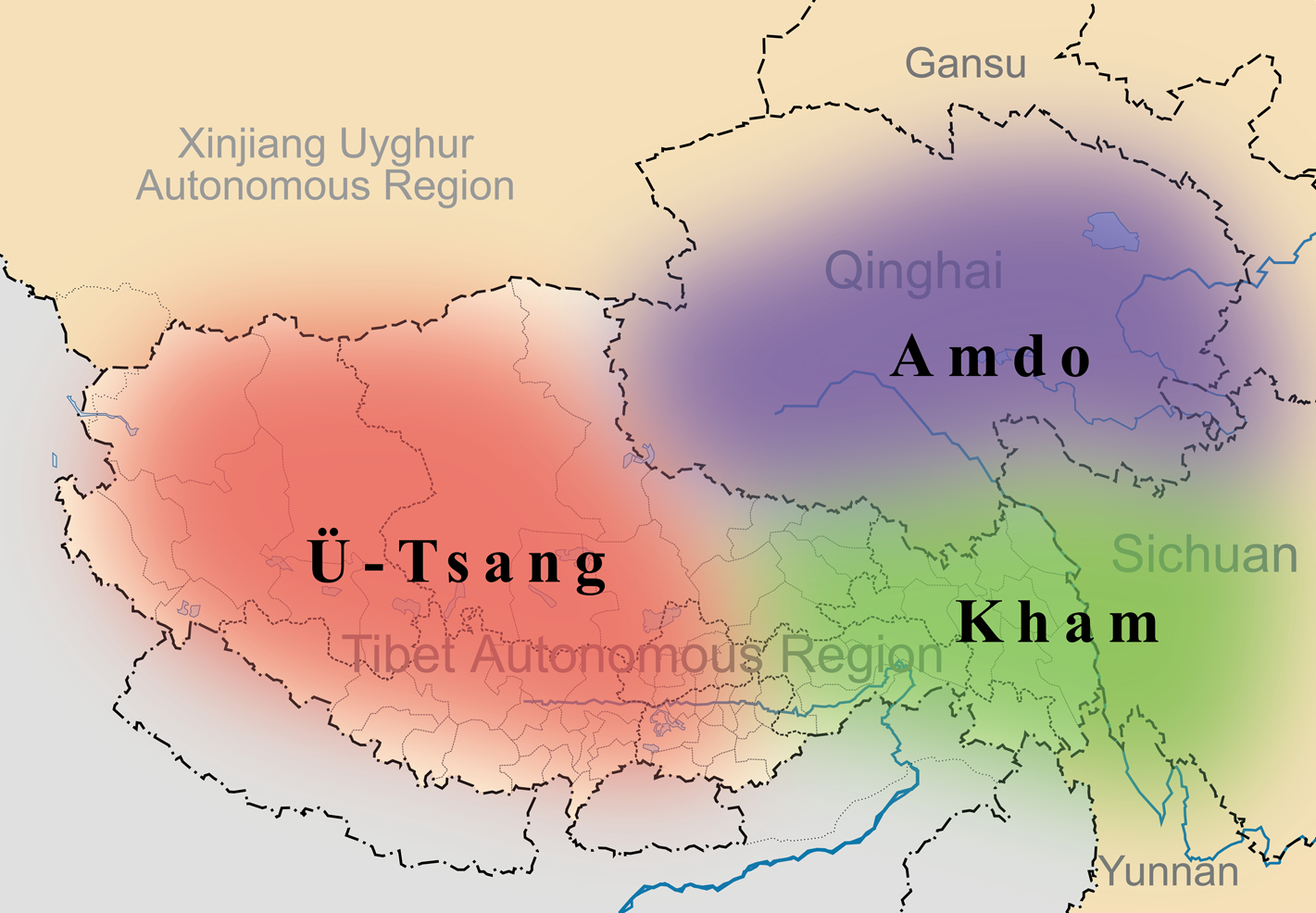}
    \caption{Approximate distribution of major Tibetan dialect groups. 
    Ü-Tsang (red), Khams (green), and Amdo (purple) show distinct 
    historical trajectories in tonogenesis.}
    \label{fig:tibetan_dialects_map}
\end{figure}

More than fifty distinct varieties have been identified as part of the Tibetic branch of the Sino-Tibetan language family \citep{tournadre_tibetic_2014}. Among these, the three most widely discussed groups are Central, Amdo, and Khams, illustrated in Figure~\ref{fig:tibetan_dialects_map} \citep{gesang_zangyu_2002}. The longstanding debate over whether these varieties constitute separate languages or dialects remains relevant \citep{haugen_dialect_1966}. From a strictly linguistic standpoint, many varieties are only marginally mutually intelligible \citep{driem_languages_2001}, suggesting that they could be treated as separate languages. Nevertheless, socio-political and cultural factors often motivate their classification as “Tibetan dialects,” highlighting a shared literary heritage (based on Classical Tibetan) and a unified Tibetan ethnic identity \citep{tournadre_tibetic_2014, gesang_zangyu_2002}. In this work, we use the term \emph{Tibetan languages}, while acknowledging the nuanced nature of the debate.

Tibetan languages exhibit a diverse range of tonogenesis outcomes. Central (Ü-Tsang) varieties such as Lhasa Tibetan have developed robust tonal distinctions from historical voicing contrasts: modern Lhasa has two phonemic tones that yield four different pitch contours depending on syllable structure \citep{delancey_lhasa_2017, lim_tonal_2018}. Khams varieties occupy an \emph{intermediate} position, relying partially on pitch while retaining residual segmental cues such as voicing contrasts or breathiness \citep{sun_variegated_2015, suzuki_preaspiration_2011}. For instance, Dege Khams preserves historically voiced consonants alongside four emerging tonal categories \citep{gesang_zangyu_2002}. At the opposite end of the spectrum, Amdo varieties remain primarily atonal, retaining a rich inventory of consonantal clusters in lieu of pitch-based contrasts \citep{gesang_zangyu_2002, sun_aspects_1986}.

\subsection{Tone and ASR}
Although pitch variations occur in all spoken languages for prosodic or paralinguistic reasons, \emph{tone languages} uniquely depend on fundamental frequency ($f_0$) as a primary lexical contrast. As a result, failing to model pitch cues in automatic speech recognition (ASR) can lead to frequent confusions among words that differ only in tone \citep{fu_importance_1998, patel_role_2010, wang_roles_2013, zhang_role_2020}. In practice, researchers typically address tone modeling in two broad ways. Some approaches add time-aligned pitch or voice-quality features directly into the acoustic model \citep{lei_improved_2006, li_improved_2011}, allowing it to learn the correlations between pitch contours and word identities. Other approaches explicitly encode tonal information in the transcript by assigning diacritics or numeric labels to otherwise identical syllables or phonemes \citep{yuan_automatic_2021, coto-solano_explicit_2021}. Regardless of the chosen method, extensive evidence shows that removing or flattening pitch severely degrades ASR performance in tonal languages, mirroring the disruption that human listeners experience with $f_0$ suppression \citep{fu_importance_1998, niu_effect_2019}. In other words, in fully (or partly) tonal systems, capturing pitch movements becomes essential for correctly mapping acoustic signals to lexical items.

Recent work has demonstrated that systematically removing pitch through flattening provides a computational method for quantifying how heavily languages rely on $f_0$ for lexical disambiguation: fully tonal systems suffer severe ASR degradation when pitch is removed, while non-tonal languages show minimal impact \citep{liang_tone_2025}. This methodology offers a way to assess the functional load of tone in ASR systems, complementing traditional linguistic measures based on minimal pair counts.

\subsection{Functional Load of Tone}
In phonetics and phonology, the \emph{functional load} of a contrast refers to how crucial that contrast is for distinguishing words \citep{surendran_functional_2004}. In a strongly tonal language such as Mandarin or Cantonese, this notion conceptually relates to the amount of minimal pairs that depend solely on tone or to gauging how much pitch contributes to speech intelligibility. High functional load corresponds to greater confusion when $f_0$ is removed or altered, which in turn leads to marked increases in ASR error rates.

However, this measure can be inadequate for tonogenesis languages, which are still shifting from consonant-based distinctions (e.g.\ voicing or glottal stop codas) to primarily pitch-based contrasts \citep{hombert_phonetic_1979, ratliff_tonoexodus_2015}. In many transitional systems, tone can remain partially predictable from segmental cues, coarticulation, or laryngeal features \citep{hu_tonogenesis_2012, sun_variegated_2015}. For example, pitch contours might be strongly influenced by the presence of voiced versus aspirated onsets \citep{hombert_development_1977}, or by residual glottalization that masks or enhances emerging pitch distinctions \citep{kirby_transphonologization_2022}. These extra cues mean that a simple tally of tone-based minimal pairs overestimates how much $f_0$ alone is responsible for contrasts. Even after $f_0$ becomes phonemic, features like breathiness or preaspiration sometimes continue to overlap with pitch distinctions, complicating the idea of an independent “tone functional load.” Consequently, although functional load remains a productive concept for mature tone languages, a single pitch-based metric does not fully capture the nuanced interplay between pitch and segmental cues in incipient or partial tone systems. In these contexts, more holistic approaches are needed to account for how tone interacts with residual voicing, glottal stops, and other acoustic features—both for phonological description and for building robust ASR models. 

\section{Methods}

We employ a pitch-flattening methodology to investigate the tonogenesis continuum in Tibetan varieties. This method allows us to quantify how heavily each dialect relies on pitch for lexical contrasts by measuring ASR performance degradation when $f_0$ information is systematically removed.

\subsection{Data}

We examine six Tibetan varieties drawn from TIBMD@MUC \citep{zhao_open_2020}, described in Table~\ref{tab:tibmdmuc}. Recordings featured both formal and colloquial read speech, recorded at 16 kHz, transcribed in Tibetan script. In each variety, we sampled approximately two hours of audio for training. We attempted to limit training data to two speakers per variety; in Changdu and Dege, however, three speakers were used due to data constraints. From each variety, we also reserved around 30 minutes of speech to form a test set. Data from Yushu were excluded due to insufficient data.

Since Tibetan orthography relies on stacked letters and diacritics, we converted the script to Wylie transliterations via \texttt{pyewts} \citep{wylie_standard_1959}. We then segmented each Wylie transcription at the character level (treating spaces as distinct symbols) to form a language-specific vocabulary. Audio files were resampled at 16 kHz, and we fine-tuned separate XLS-R 300m models \citep{babu_xls-r_2021} using standard hyperparameters for low-resource ASR.

\begin{table}[htbp]
\centering
\begin{tabular}{lccc}
\hline
\textbf{Group} & \textbf{\begin{tabular}[c]{@{}c@{}}Duration\\(hrs)\end{tabular}} & \textbf{Spkr} & \textbf{Utterances} \\
\hline
\multicolumn{4}{l}{\textbf{Amdo}} \\
\hline
Xiahe      & 4.12  &  2  & 3549 \\
Aba        & 8.16  &  2  & 6546 \\
Qinghai    & 13.65 & 19  & 12859 \\
\hline
\multicolumn{4}{l}{\textbf{Kham}} \\
\hline
Changdu    & 2.79  &  7  & 2558 \\
Dege       & 2.31  &  3  & 1245 \\
Yushu      & 0.77  &  3  & 631  \\
\hline
\multicolumn{4}{l}{\textbf{Ü-Tsang}} \\
\hline
Lhasa      & 37.38 & 48  & 30349 \\
Shigatse   & 15.15 &  4  & 10729 \\
\hline
\multicolumn{1}{l}{\textbf{Total}} & 84.33 & 88 & 68466 \\
\hline
\end{tabular}
\caption{Statistics of the TIBMD@MUC database, illustrating eight Tibetan languages grouped into Amdo, Kham, and Ü-Tsang.}
\label{tab:tibmdmuc}
\end{table}

\subsection{Model Training and Evaluation}

We fine-tuned individual XLS-R 300m models for each of the six Tibetan varieties. During preprocessing, we normalized text and generated a character-based vocabulary for CTC training. Following model convergence, we evaluated each model on both the original test set and a pitch-flattened version of the same data.

Pitch flattening was performed using Praat's Pitch-Synchronous OverLap and Add (PSOLA) algorithm, which replaces each utterance's natural $f_0$ contour with its mean pitch. This manipulation effectively removes lexical tone cues while preserving the spectral envelope and temporal structure of the speech signal. By comparing character error rate (CER) and word error rate (WER) before and after flattening, we obtain a measure of how much each variety's ASR system relies on pitch for lexical disambiguation.

Because Ü-Tsang varieties (e.g., Lhasa) are further along the tonogenesis trajectory, we hypothesized they would suffer greater performance drops upon pitch removal, reflecting a heavier functional load for $f_0$ than in the more atonal Amdo dialects. Kham varieties, described linguistically as occupying an intermediate position, were expected to show moderate sensitivity to pitch manipulation.

\section{Results}

Table~\ref{tab:tibetan_results} presents ASR outcomes for six Tibetan languages under original vs.\ pitch-flattened conditions, revealing clear differences in how each variety responds to the removal of $f_0$. Amdo dialects (Xiahe, Aba), generally described as \emph{atonal} or pre-tonogenetic \citep{sun_aspects_1986, gesang_zangyu_2002}, exhibit only moderate increases in error rates after pitch flattening, suggesting that segmental contrasts (e.g., voiced vs.\ voiceless onsets) continue to carry the primary lexical burden. By contrast, the Ü-Tsang group (Lhasa, Shigatse), which has been characterized as fully tonal \citep{delancey_lhasa_2017}, incurs notably larger performance drops. Shigatse in particular shows a sharp spike in CER (+0.139), indicating that obscuring pitch removes a principal mechanism of word-level contrast, much as in other \emph{mature} tone systems \citep{lim_tonal_2018, kirby_incipient_2014}.

Falling between these two poles are the Kham varieties (Changdu, Dege), which illustrate a partially developed reliance on pitch. Changdu undergoes a moderate rise in CER and WER when pitch is flattened, implying that $f_0$ is emerging as a contrastive feature yet does not fully supersede residual segmental or phonation-based cues \citep{sun_variegated_2015}. Dege, despite having a higher baseline error rate overall, exhibits comparatively small deltas under pitch removal, suggesting that its lexical contrasts may still be anchored in non-pitch features, or that limited training data has led to underrepresentation of the language's tonal cues. Together, these outcomes corroborate the view that Kham varieties have not completed the shift toward fully developed tone systems \citep{suzuki_preaspiration_2011, kirby_transphonologization_2022}.

Overall, this pattern of results supports the broader idea of a \emph{tonogenesis continuum} in Tibetan: atonal Amdo dialects appear minimally impacted by pitch manipulation, Ü-Tsang's well-established tones produce sizable recognition deficits, and Kham falls in between, reflecting an incomplete transition away from historical voicing contrasts toward $f_0$-based distinctions. These findings underscore the importance of capturing both segmental and suprasegmental factors in Tibetan ASR. They also suggest that, as a dialect shifts deeper into tonogenesis, the stakes of losing pitch cues rise commensurately, culminating in fully tonal systems whose lexical contrasts become acutely vulnerable to $f_0$ flattening.

\begin{table*}[htbp]
\centering
\begin{tabular}{llcccccc}
\hline
 &  & \multicolumn{2}{c}{\textbf{Original}} & \multicolumn{2}{c}{\textbf{Flattened}} & \multicolumn{2}{c}{\textbf{$\Delta$}} \\
\cline{3-4} \cline{5-6} \cline{7-8}
\textbf{Language} & \textbf{Tone Status} & \textbf{CER} & \textbf{WER} & \textbf{CER} & \textbf{WER} & \textbf{CER} & \textbf{WER} \\
\hline
\multicolumn{8}{l}{\textbf{Amdo}} \\
\hline
Xiahe    & Non-tonal & 0.114 & 0.320 & 0.139 & 0.378 & 0.025 & 0.058 \\
Aba      & Non-tonal & 0.182 & 0.525 & 0.202 & 0.563 & 0.020 & 0.038 \\
\hline
\multicolumn{8}{l}{\textbf{Ü-Tsang}} \\
\hline
Lhasa    & Tonal     & 0.177 & 0.486 & 0.237 & 0.593 & 0.060 & 0.107 \\
Shigatse & Tonal     & 0.490 & 0.175 & 0.629 & 0.250 & 0.139 & 0.075 \\
\hline
\multicolumn{8}{l}{\textbf{Kham}} \\
\hline
Changdu  & Tonal     & 0.247 & 0.523 & 0.303 & 0.613 & 0.056 & 0.090 \\
Dege     & Tonal     & 0.475 & 0.902 & 0.492 & 0.917 & 0.017 & 0.015 \\
\hline
\end{tabular}
\caption{Character error rate (CER) and word error rate (WER) for six Tibetan dialects under 
original vs.\ pitch-flattened conditions, along with their respective $\Delta$ (Flattened--Original). 
Dialects are grouped by tonal status.}
\label{tab:tibetan_results}
\end{table*}

\section{Discussion}
\label{sec:discussion}

Our experiments offer insight into how pitch-flattening impacts ASR across Tibetan dialects at different stages of tonogenesis. Table~\ref{tab:tibetan_results} reveals a clear gradient pattern: the atonal Amdo varieties (Xiahe, Aba) experience only small increases in CER and WER upon pitch-flattening, pointing to a reliance on residual consonantal cues (e.g.\ complex onset clusters, voicing) rather than $f_0$ \citep{gesang_zangyu_2002, sun_aspects_1986}. At the other end of the tonogenesis continuum, the fully tonal Ü-Tsang dialects (Lhasa, Shigatse) incur larger recognition penalties, indicating that pitch has become a primary mechanism for lexical contrast \citep{delancey_lhasa_2017, lim_tonal_2018}.

The Kham varieties (Changdu, Dege) occupy an \emph{intermediate} position consistent with descriptions of partial tonogenesis in which pitch has begun to assume a contrastive load but still coexists with other cues \citep{sun_variegated_2015, suzuki_preaspiration_2011}. Changdu exhibits moderate performance drops (+0.056 CER, +0.090 WER), suggesting that both pitch and residual segmental features remain important. Dege shows a smaller delta despite a higher overall error rate, possibly indicating that the dialect's emergent tonal distinctions are underpinned by breathiness or voicing contrasts, or that limited training data dampened the model's reliance on $f_0$ cues. Taken together, these patterns strengthen the conclusion that Tibetan dialects vary in pitch dependence according to their historical stage of tonogenesis, mirroring prior research highlighting the gradual shift from archaic voicing contrasts to full-blown pitch-based systems \citep{haudricourt_lorigine_1954, hombert_phonetic_1979, ratliff_tonoexodus_2015}.

An interesting fact of Tibetan languages is the use of a shared script dating back to Classical Tibetan. This orthography encodes older consonant clusters and voicing distinctions, thus leaving tonal contrasts unmarked for modern varieties as a result of diachronic development \citep{wylie_standard_1959}. Consequently, the ASR models must learn any pitch-based distinctions directly from acoustic input. For dialects like Amdo, which retain many of the historical segmental cues, pitch contributes little to lexical disambiguation and can be removed with minimal impact on recognition. In fully tonal dialects like Lhasa, however, sound change has reduced or eliminated these consonantal cues, rendering $f_0$ indispensable for word identification \citep{lim_tonal_2018}.

From a typological standpoint, these results shed fresh light on how functional load interacts with partial or incipient tone. The notion of \emph{functional load} proposes that the more minimal pairs rely on a given contrast, the more catastrophic it is to remove that contrast from the signal \citep{surendran_functional_2004}. However, in a transitional system such as Kham, pitch may be correlated with, or redundant to, residual voicing or breathiness cues \citep{sun_variegated_2015, brunelle_tone_2016}. As a result, simply counting tone-based minimal pairs can overstate how often pitch alone distinguishes lexical items \citep{hombert_phonetic_1979, ratliff_tonoexodus_2015}. Our work shows that flattening $f_0$ does degrade recognition in Kham, but not universally to the extent observed in fully tonal dialects. Thus, while functional load captures \emph{some} of the story, it cannot fully explain how segmental and suprasegmental cues get dynamically reweighted throughout tonogenesis. A purely text-based estimate of tone's load would miss the subtle interplay between pitch and other laryngeal features in transitional dialects \citep{hu_tonogenesis_2012, kirby_transphonologization_2022}.

Overall, these findings highlight the practical challenge for Tibetan ASR of capturing both the legacy of archaic consonant distinctions and the emergent or fully developed reliance on pitch. As \citet{zhao_open_2020, qin_improving_2022} have shown, new acoustic features such as voice quality and breathiness may improve robustness for dialects in flux. A next step is a more granular error analysis, focusing on whether certain segments or syllables are more prone to confusion when $f_0$ is removed. Such studies could clarify how partial tone systems, where voicing and pitch contrasts remain intertwined, may ultimately evolve into the strongly pitch-dependent systems exemplified by Lhasa. In addition, these insights could motivate new approaches to multi-dialect Tibetan ASR, perhaps by selectively conditioning on pitch or phonation cues only where they carry crucial lexical information.

In sum, our experiments confirm that tone systems operate along a broad continuum, from atonal to partly tonal to fully tonal. By mapping each Tibetan language's performance profile under pitch-flattening, we obtain evidence for how the process of tonogenesis reshapes a language's reliance on $f_0$. The results not only corroborate long-standing theories of tone development \citep{matisoff_tibeto-burman_1999, hombert_phonetic_1979, haudricourt_lorigine_1954}, but also point to new computational challenges. For emerging tone systems, we must move beyond static estimates of functional load to account for how segmental and suprasegmental cues blend to encode lexical meaning.

\begin{figure}[t]
    \centering
    \includegraphics[width=4cm]{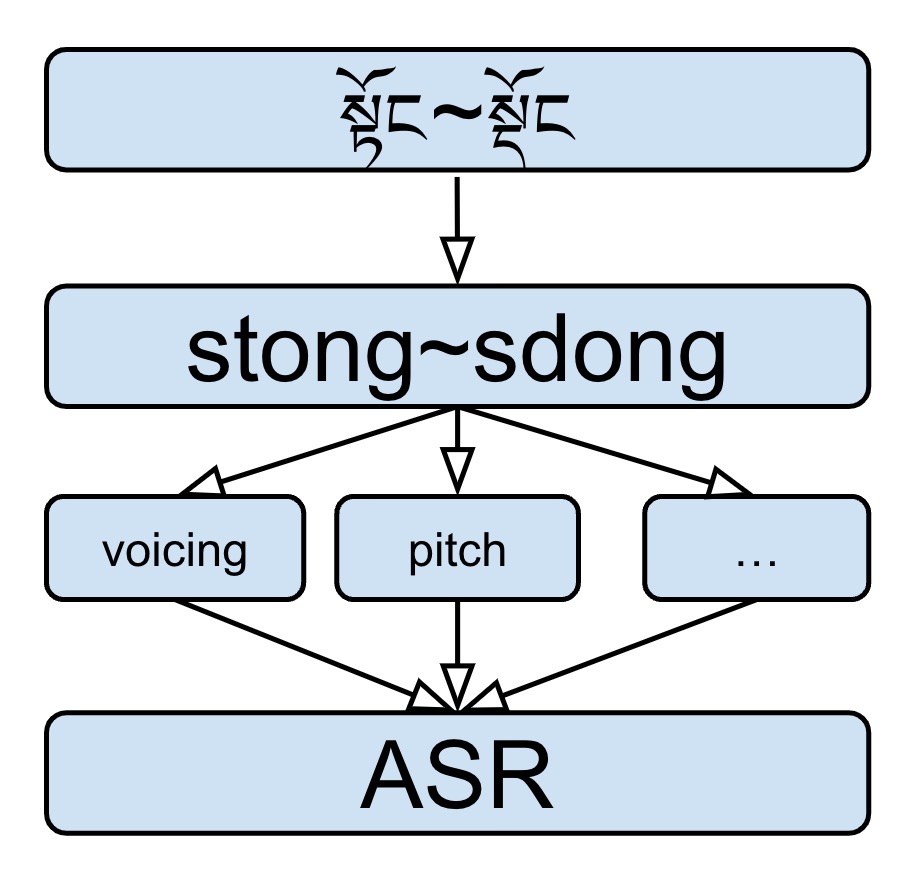}
    \caption{A schematic illustrating how Tibetan script maps to Wylie, then to acoustic features such as pitch and voicing, which the ASR model weighs during decoding.}
    \label{fig:diagram}
\end{figure}

\section{Conclusion}

We have shown that pitch-flattening reveals systematic variation in ASR performance across Tibetan varieties that directly corresponds to their position along the tonogenesis continuum. Amdo dialects (e.g., Xiahe, Aba) incurred minimal penalty from pitch suppression, reflecting their atonal nature and continued reliance on consonantal contrasts. Ü-Tsang dialects (Lhasa, Shigatse) suffered sharp error increases that underscore the functional load of pitch for lexical distinctions in fully tonal systems. Kham varieties (Changdu, Dege) fell between these extremes, showing partial reliance on $f_0$ alongside residual voicing or phonation cues.

From an ASR standpoint, these findings reinforce that a one-size-fits-all acoustic modeling approach may be insufficient for languages undergoing or nearing completion of tonogenesis. Systems trained on orthographies that only encode historical clusters or voicing may inadvertently overlook emergent pitch contrasts, or conversely fail to leverage lingering segmental cues in transitional dialects. Our results suggest that models can achieve more robust performance by incorporating explicit pitch or voice-quality features and by dynamically reweighting cues based on each dialect's stage of tonogenesis. In turn, these insights open the door for improved multi-dialect systems that capture both legacy consonantal distinctions and newly phonologized tone categories.

Overall, our work highlights how pitch manipulation offers a straightforward window on the evolving functional load of tone. While techniques like measuring minimal pairs or examining segmental environments can provide a textual or phonological perspective on tone, computational experiments using pitch-flattening yield concrete evidence of how critical $f_0$ is for modern-day lexical discrimination. In doing so, we situate Tibetan within a cross-linguistic tonogenetic trajectory that further underscores the interplay of segmental and suprasegmental features. Future research could extend these experiments by incorporating more dialects in flux, examining detailed confusion patterns for specific segments, or exploring advanced acoustic correlates (e.g., breathiness) that may serve as additional or alternative cues in emergent tone systems.

\section{Limitations}

Several limitations affect the scope and generalizability of our findings. First, our study examines only six Tibetan varieties, which, while representing key points on the tonogenesis continuum, cannot capture the full diversity of Tibetan dialectology. The TIBMD@MUC corpus provides valuable data but is constrained by uneven speaker representation across varieties and relatively small test sets (approximately 30 minutes per dialect).

Second, due to the realities of linguistic fieldwork, our training and test datasets often contained recordings from the same speakers. This may inflate performance estimates and limit our ability to assess model robustness to speaker variability. Future work should validate these findings with genuinely held-out speakers to better gauge generalization.

Third, the shared Tibetan orthography, while linguistically interesting, presents challenges for ASR evaluation. The script encodes historical consonant clusters rather than modern phonological contrasts, and transcription conventions may vary within and across datasets. This inconsistency particularly affects our ability to assess errors in phonologically complex categories.

\bibliography{custom}

\begin{thebibliography}{38}
\providecommand{\natexlab}[1]{#1}

\bibitem[{Babu et~al.(2021)Babu, Wang, Tjandra, Lakhotia, Xu, Goyal, Singh, von
  Platen, Saraf, Pino, Baevski, Conneau, and Auli}]{babu_xls-r_2021}
Arun Babu, Changhan Wang, Andros Tjandra, Kushal Lakhotia, Qiantong Xu, Naman
  Goyal, Kritika Singh, Patrick von Platen, Yatharth Saraf, Juan Pino, Alexei
  Baevski, Alexis Conneau, and Michael Auli. 2021.
\newblock \href {https://doi.org/10.48550/arXiv.2111.09296} {{XLS}-{R}:
  {Self}-supervised {Cross}-lingual {Speech} {Representation} {Learning} at
  {Scale}}.
\newblock \emph{arXiv preprint}.
\newblock ArXiv:2111.09296 [cs, eess].

\bibitem[{Brunelle and Kirby(2016)}]{brunelle_tone_2016}
Marc Brunelle and James Kirby. 2016.
\newblock \href {https://doi.org/10.1111/lnc3.12182} {Tone and {Phonation} in
  {Southeast} {Asian} {Languages}: {Tone} and {Phonation} in {Southeast}
  {Asian} {Languages}}.
\newblock \emph{Language and Linguistics Compass}, 10(4):191--207.

\bibitem[{Coto-Solano(2021)}]{coto-solano_explicit_2021}
Rolando Coto-Solano. 2021.
\newblock \href {https://doi.org/10.18653/v1/2021.americasnlp-1.20} {Explicit
  {Tone} {Transcription} {Improves} {ASR} {Performance} in {Extremely}
  {Low}-{Resource} {Languages}: {A} {Case} {Study} in {Bribri}}.
\newblock In \emph{Proceedings of the {First} {Workshop} on {Natural}
  {Language} {Processing} for {Indigenous} {Languages} of the {Americas}},
  pages 173--184, Online. Association for Computational Linguistics.

\bibitem[{DeLancey(2017)}]{delancey_lhasa_2017}
Scott DeLancey. 2017.
\newblock Lhasa {Tibetan}.
\newblock In \emph{The {Sino}-{Tibetan} {Languages}}.

\bibitem[{Driem(2001)}]{driem_languages_2001}
George~van Driem. 2001.
\newblock \href {https://brill.com/display/title/17519} {Languages of the
  {Himalayas}}.
\newblock Brill.

\bibitem[{Edwards and Beckman(1988)}]{edwards_articulatory_1988}
Jan Edwards and Mary~E. Beckman. 1988.
\newblock \href {https://doi.org/10.1159/000261824} {Articulatory {Timing} and
  the {Prosodic} {Interpretation} of {Syllable} {Duration}}.
\newblock \emph{Phonetica}, 45(2-4):156--174.
\newblock Publisher: De Gruyter Mouton.

\bibitem[{Fu et~al.(1998)Fu, Zeng, Shannon, and Soli}]{fu_importance_1998}
Qian-Jie Fu, Fan-Gang Zeng, Robert~V Shannon, and Sigfrid~D Soli. 1998.
\newblock Importance of tonal envelope cues in {Chinese} speech recognition.
\newblock \emph{The Journal of the Acoustical Society of America},
  104(1):505--510.
\newblock Publisher: Acoustical Society of America.

\bibitem[{Gesang and Gesang(2002)}]{gesang_zangyu_2002}
Jumian Gesang and Yangjing Gesang. 2002.
\newblock Zangyu fangyan gailun.
\newblock \emph{Beijing: Minzu chubanshe}.

\bibitem[{Haudricourt(1954)}]{haudricourt_lorigine_1954}
André-Georges Haudricourt. 1954.
\newblock De l’origine des tons en vietnamien.
\newblock \emph{Journal Asiatique}, 242:69--82.

\bibitem[{Haugen(1966)}]{haugen_dialect_1966}
Einar Haugen. 1966.
\newblock \href {https://doi.org/10.1525/aa.1966.68.4.02a00040} {Dialect,
  {Language}, {Nation}}.
\newblock \emph{American Anthropologist}, 68(4):922--935.

\bibitem[{Hombert(1977)}]{hombert_development_1977}
Jean-Marie Hombert. 1977.
\newblock \href {https://doi.org/10.1016/S0095-4470(19)31109-X} {Development of
  tones from vowel height?}
\newblock \emph{Journal of Phonetics}, 5(1):9--16.

\bibitem[{Hombert et~al.(1979)Hombert, Ohala, and Ewan}]{hombert_phonetic_1979}
Jean-Marie Hombert, John~J. Ohala, and William~G. Ewan. 1979.
\newblock \href {https://doi.org/10.2307/412518} {Phonetic {Explanations} for
  the {Development} of {Tones}}.
\newblock \emph{Language}, 55(1):37--58.
\newblock Publisher: Linguistic Society of America.

\bibitem[{Hu(2012)}]{hu_tonogenesis_2012}
Fang Hu. 2012.
\newblock Tonogenesis in {Lhasa} {Tibetan} – {Towards} a gestural account.
\newblock \emph{Consonant Clusters and Structural Complexity}, 26:231.

\bibitem[{Kirby et~al.(2022)Kirby, Pittayaporn, and
  Brunelle}]{kirby_transphonologization_2022}
James Kirby, Pittayawat Pittayaporn, and Marc Brunelle. 2022.
\newblock \href {https://doi.org/10.1515/phon-2022-0029} {Transphonologization
  of onset voicing: revisiting {Northern} and {Eastern} {Kmhmu}’}.
\newblock \emph{Phonetica}, 79(6):591--629.

\bibitem[{Kirby(2014)}]{kirby_incipient_2014}
James~P. Kirby. 2014.
\newblock \href {https://doi.org/10.1016/j.wocn.2014.02.001} {Incipient
  tonogenesis in {Phnom} {Penh} {Khmer}: {Acoustic} and perceptual studies}.
\newblock \emph{Journal of Phonetics}, 43:69--85.

\bibitem[{Lei et~al.(2006)Lei, Siu, Hwang, Ostendorf, and
  Lee}]{lei_improved_2006}
Xin Lei, Manhung Siu, Mei-Yuh Hwang, Mari Ostendorf, and Tan Lee. 2006.
\newblock \href {https://doi.org/10.21437/Interspeech.2006-372} {Improved tone
  modeling for {Mandarin} broadcast news speech recognition}.
\newblock In \emph{Interspeech 2006}, pages paper 1752--Tue3A2O.4--0. ISCA.

\bibitem[{Li et~al.(2011)Li, Wang, Sun, and Lee}]{li_improved_2011}
Shang-wen Li, Yow-bang Wang, Liang-che Sun, and Lin-shan Lee. 2011.
\newblock \href {https://doi.org/10.21437/Interspeech.2011-609} {Improved tonal
  language speech recognition by integrating spectro-temporal evidence and
  pitch information with properly chosen tonal acoustic units}.
\newblock In \emph{Interspeech 2011}, pages 2293--2296. ISCA.

\bibitem[{Liang and Levow(2025)}]{liang_tone_2025}
Siyu Liang and Gina-Anne Levow. 2025.
\newblock \href {https://doi.org/10.18653/v1/2025.sigtyp-1.11} {Tone in
  {Perspective}: {A} {Computational} {Typological} {Analysis} of {Tone}
  {Function} in {ASR}}.
\newblock In \emph{Proceedings of the 7th {Workshop} on {Research} in
  {Computational} {Linguistic} {Typology} and {Multilingual} {NLP}}, pages
  82--92, Vienna, Austria. Association for Computational Linguistics.

\bibitem[{Lim(2018)}]{lim_tonal_2018}
Keh~Sheng Lim. 2018.
\newblock \emph{The {Tonal} and {Intonational} {Phonology} of {Lhasa}
  {Tibetan}}.
\newblock Ph.D. thesis.

\bibitem[{Matisoff(1973)}]{matisoff_tonogenesis_1973}
James~A Matisoff. 1973.
\newblock Tonogenesis in southeast {Asia}.
\newblock \emph{Consonant types and tone}, 1(1):71--96.

\bibitem[{Matisoff(1999)}]{matisoff_tibeto-burman_1999}
James~A Matisoff. 1999.
\newblock Tibeto-{Burman} tonology in an areal context.
\newblock In \emph{Proceedings of the symposium “{Crosslinguistic} studies of
  tonal phenomena: {Tonogenesis}, {Japanese} {Accentology}, and {Other}
  {Topics}}, pages 3--31. Tokyo: Tokyo University of Foreign Studies, Institute
  for the Study of ….

\bibitem[{Matthews and Yip(2013)}]{matthews_cantonese_2013}
Stephen Matthews and Virginia Yip. 2013.
\newblock \href {https://doi.org/10.4324/9780203835012} {\emph{Cantonese: {A}
  {Comprehensive} {Grammar}}}, 2 edition.
\newblock Routledge, London.

\bibitem[{Niu et~al.(2019)Niu, Chen, and Chen}]{niu_effect_2019}
Yadong Niu, Fei Chen, and Jing Chen. 2019.
\newblock \href {https://doi.org/10.1121/1.5119264} {The effect of {F0} contour
  on the intelligibility of {Mandarin} {Chinese} for hearing-impaired
  listeners}.
\newblock \emph{The Journal of the Acoustical Society of America},
  146(2):EL85--EL91.

\bibitem[{Patel et~al.(2010)Patel, Xu, and Wang}]{patel_role_2010}
Aniruddh~D. Patel, Yi~Xu, and Bei Wang. 2010.
\newblock \href {https://doi.org/10.21437/SpeechProsody.2010-238} {The role of
  {F0} variation in the intelligibility of {Mandarin} sentences}.
\newblock In \emph{Speech {Prosody} 2010}, pages paper 890--0. ISCA.

\bibitem[{Qin et~al.(2022)Qin, Wang, Li, Dang, and Pan}]{qin_improving_2022}
Siqing Qin, Longbiao Wang, Sheng Li, Jianwu Dang, and Lixin Pan. 2022.
\newblock \href {https://doi.org/10.1186/s13636-021-00233-4} {Improving
  low-resource {Tibetan} end-to-end {ASR} by multilingual and multilevel unit
  modeling}.
\newblock \emph{EURASIP Journal on Audio, Speech, and Music Processing},
  2022(1):2.

\bibitem[{Ratliff(2015)}]{ratliff_tonoexodus_2015}
Martha Ratliff. 2015.
\newblock \href {https://doi.org/10.1093/oxfordhb/9780199232819.013.021}
  {Tonoexodus, {Tonogenesis}, and {Tone} {Change}}.
\newblock In Patrick Honeybone and Joseph Salmons, editors, \emph{The {Oxford}
  {Handbook} of {Historical} {Phonology}}, page~0. Oxford University Press.

\bibitem[{Sun(2015)}]{sun_variegated_2015}
Jackson (Ed~) Sun. 2015.
\newblock \href {https://doi.org/10.15144/PL-555.35} {Variegated tonal
  developments in {Tibetan}}.
\newblock pages 2.1M, 35--52 pages.
\newblock Artwork Size: 2.1M, 35-52 pages Medium: PDF Publisher: Pacific
  Linguistics Version Number: 1.0.

\bibitem[{Sun(1986)}]{sun_aspects_1986}
Jackson T-S Sun. 1986.
\newblock Aspects of the {Phonology} of {Amdo} {Tibetan}.
\newblock \emph{(No Title)}.

\bibitem[{Surendran and Levow(2004)}]{surendran_functional_2004}
Dinoj Surendran and Gina-Anne Levow. 2004.
\newblock \href {https://doi.org/10.21437/SpeechProsody.2004-23} {The
  functional load of tone in {Mandarin} is as high as that of vowels}.
\newblock In \emph{Speech {Prosody} 2004}, pages 99--102. ISCA.

\bibitem[{Suzuki(2011)}]{suzuki_preaspiration_2011}
Hiroyuki Suzuki. 2011.
\newblock Preaspiration and tonal development in {Tibetan} dialects of {Khams},
  {Shar} and {Amdo}.
\newblock In \emph{Tone, accent and intonation in eastern {Eurasian} languages.
  {The} 18th {Meeting} of the {Linguistic} {Circle} for the {Study} of
  {Eastern} {Eurasian} {Languages}, {Aoyama} {Gakuin} {University}, {Tokyo}},
  pages 9--17.

\bibitem[{Thurgood(2002)}]{thurgood_vietnamese_2002}
Graham Thurgood. 2002.
\newblock \href {https://doi.org/10.1075/dia.19.2.04thu} {Vietnamese and
  tonogenesis: {Revising} the model and the analysis}.
\newblock \emph{Diachronica}, 19(2):333--363.

\bibitem[{Tournadre(2014)}]{tournadre_tibetic_2014}
Nicolas Tournadre. 2014.
\newblock The {Tibetic} languages and their classification.
\newblock \emph{Trans-Himalayan linguistics: Historical and descriptive
  linguistics of the Himalayan area}, 266(1):105--29.

\bibitem[{Wang et~al.(2013)Wang, Shu, Zhang, Liu, and Zhang}]{wang_roles_2013}
Jiuju Wang, Hua Shu, Linjun Zhang, Zhaoxing Liu, and Yang Zhang. 2013.
\newblock \href {https://doi.org/10.1121/1.4811159} {The roles of fundamental
  frequency contours and sentence context in {Mandarin} {Chinese} speech
  intelligibility}.
\newblock \emph{The Journal of the Acoustical Society of America},
  134(1):EL91--EL97.

\bibitem[{Wylie(1959)}]{wylie_standard_1959}
Turrell Wylie. 1959.
\newblock A standard system of {Tibetan} transcription.
\newblock \emph{Harvard journal of Asiatic studies}, 22:261--267.
\newblock Publisher: JSTOR.

\bibitem[{Yip(2002)}]{yip_tone_2002}
Moira Jean~Winsland Yip. 2002.
\newblock \emph{Tone}.
\newblock Cambridge textbooks in linguistics. Cambridge University Press,
  Cambridge ; New York.

\bibitem[{Yuan et~al.(2021)Yuan, Ryant, Cai, Church, and
  Liberman}]{yuan_automatic_2021}
Jiahong Yuan, Neville Ryant, Xingyu Cai, Kenneth Church, and Mark Liberman.
  2021.
\newblock \href {https://doi.org/10.48550/arXiv.2108.01122} {Automatic
  recognition of suprasegmentals in speech}.
\newblock \emph{arXiv preprint}.
\newblock ArXiv:2108.01122 [cs].

\bibitem[{Zhang and Kirby(2020)}]{zhang_role_2020}
Yubin Zhang and James Kirby. 2020.
\newblock \href {https://doi.org/10.1121/10.0001523} {The role of {F0} and
  phonation cues in {Cantonese} low tone perception}.
\newblock \emph{The Journal of the Acoustical Society of America},
  148(1):EL40--EL45.

\bibitem[{Zhao et~al.(2020)Zhao, Xu, Yue, Song, Li, Wu, and
  Ji}]{zhao_open_2020}
Yue Zhao, Xiaona Xu, Jianjian Yue, Wei Song, Xiali Li, Licheng Wu, and Qiang
  Ji. 2020.
\newblock \href {https://doi.org/10.1504/ijcse.2020.107351} {An open speech
  resource for {Tibetan} multi-dialect and multitask recognition}.
\newblock \emph{Int. J. Comput. Sci. Eng.}, 22(2-3):297--304.

\end{thebibliography}

\appendix
\section{Appendix}
\label{sec:appendix}

This appendix provides additional details on our fine-tuning hyperparameters for XLS-R 300m in both experiments.

\subsection{XLS-R Fine-Tuning Hyperparameters}
All training runs (for both Common Voice and TIBMD@MUC data) used the same set of essential hyperparameters, with only minor adjustments for batch size depending on GPU memory:
\begin{itemize}
    \item \textbf{Model:} \texttt{facebook/wav2vec2-xls-r-300m}
    \item \textbf{CTC loss reduction:} \emph{mean}
    \item \textbf{Batch Size:} 4 or 8 per device (gradient accumulation steps adjusted to keep effective batch size at 16)
    \item \textbf{Learning Rate:} \(3 \times 10^{-4}\)
    \item \textbf{Optimizer:} AdamW with \(\beta_1=0.9\), \(\beta_2=0.999\)
    \item \textbf{Warmup Steps:} 500
    \item \textbf{Max Steps:} 2000
    \item \textbf{Vocabulary Size:} based on unique characters in the training corpus (including space or \texttt{|} as word delimiter).
\end{itemize}

\end{document}